# Paired Comparisons-based Interactive Differential Evolution

Hideyuki Takagi
Kyushu University
Faculty of Design
Fukuoka, Japan
http://www.design.kyushu-u.ac.jp/~takagi

Denis Pallez
Nice Sophia-Antipolis University
I3S Laboratory – CNRS UMR 6070
Nice, France
http://www.i3s.unice.fr/~dpallez/

*Abstract*—We propose Interactive Differential Evolution (IDE) based on paired comparisons for reducing user fatigue and evaluate its convergence speed in comparison with Interactive Genetic Algorithms (IGA) and tournament IGA. User interface and convergence performance are two big keys for reducing Interactive Evolutionary Computation (IEC) user fatigue. Unlike IGA and conventional IDE, users of the proposed IDE and tournament IGA do not need to compare whole individuals each other but compare pairs of individuals, which largely decreases user fatigue. In this paper, we design a pseudo-IEC user and evaluate another factor, IEC convergence performance, using IEC simulators and show that our proposed IDE converges significantly faster than IGA and tournament IGA, i.e. our proposed one is superior to others from both user interface and convergence performance points of view.

*Evolutionary Algorithms; Differential Evolution; Interactive Evolutionary Computation, Paired Comparison, Gaussian Mixture Model*

## I. INTRODUCTION

There are many optimization tasks that it is not easy or almost impossible to design scales for evaluating target systems quantitatively but that we can evaluate subjectively. Some of these tasks include, for example, drawing montage based on witness's memory, fitting a hearing-aid to get satisfactory sounds, designing cute or lovely motions of home robots. Interactive Evolutionary Computation (IEC) has been applied to these tasks in a wide variety of application areas [1].

The biggest drawback of the IEC is IEC user fatigue due to human cooperation with tireless computer. IEC user has to evaluate generated individuals, which makes the user boring and tired. The population size and evolving generations are limited due to the fatigue, and 10 - 20 individuals and 10 - 20 generations are frequently used in IEC, but they are quite fewer than those of normal EC search and result slower convergence. The slower convergence is other factor of the IEC user fatigue.

Several trials have been done to solve the fatigue problem [1]. Improving IEC user interface is one of them. Some of them are: improving display interface to help user to compare individuals easily, improving input interface by reducing evaluation levels, for example 5 evaluation levels rather than 100 ones, and data visualization by projecting the distribution of individuals in the $n$-D searching space onto 2-D or 3-D space, and others .

Predicting IEC user's evaluation using trained prediction models is other trial. The model is used as a fitness function of normal EC search and is combined with IEC to accelerate IEC search. To train these models using IEC user's evaluations in past generations, a distance-based model [1], genetic programming [2], neural networks [1], and Support Vector Machine [3, 4] has been used. References [5, 6] propose to use an eye-tracker to avoid the input of fitness values and even to avoid the user for selecting best individuals: it could be done by correctly interpreting cognitive store data; for instance, by considering time spent on evaluating an individual.

Another possible solution is to use tournament IEC [7] that is based on only paired comparisons rather than comparing all individuals. This technique is easier for an IEC user to evaluate a pair of individuals than to compare all individuals and evaluate them at once. However, drawback is its less information for giving fitness values to all individuals due to lack of comparison of all individuals, which means that the fitness includes more noise and may result slow convergence. Simulated breeding is an IEC method that IEC user just chooses better individuals among all, and one click selection of a pointing device is an easy IEC user interface. Although it compares all individuals unlike tournament IEC, the 1 bit evaluation includes more quantization noise in fitness than any other evaluations and may makes IEC convergence slower [8].

Introducing new type of EC and accelerating IEC convergence are other approaches. Differential Evolution (DE) [9] is an optimization technique come to be frequently used in this decade. DE has two possible advantages that completely fall into IEC conditions: first one is the use of comparisons between only two individuals (*paired comparisons*) and second one is its potential of faster convergence. The objective of this paper is to evaluate Interactive DE (IDE) [10-12] in comparison with conventional IEC approaches and show its potential.

Unfortunately, IDE in the references [10-12] did not use the first feature of paired comparisons. Particle Swarm Optimization (PSO) can be used as an EC part in IEC. As PSO is sensitive to quantization noise in IEC fitness, better performance of Interactive PSO than Interactive Genetic Algorithms (IGA: Interactive GA) is achieved by combining some methods reducing the effect of the quantization noise in fitness with Interactive PSO [8].

The objective of this paper is to show better convergence performance of our proposed paired comparison-based IDE than convectional IEC algorithms through IEC simulation. Since the advantage of its paired comparison in comparison with many comparison of conventional IGA is obvious, the proposed IDE is the best if the proposed IDE is faster than tournament IGA or faster than or equal to IGA.

We explain EC algorithms used in our study including GA and DE in section II and how individuals are evaluated in interactive frameworks in section III. Section IV evaluates how IDE converges in comparison of three conventional IEC approaches.

## II. EC ALGORITHMS

We compare four EC algorithms (DE, Genetic Algorithm (GA), tournament1-GA, and tournament2-GA) with/without an IEC framework. Let us first present what is tournament-GA and next what is DE.

### A. Tournament Genetic Algorithms (TGA)

Reference [13] was the first who proposed *competitive fitness* that does not use absolute values of a fitness function but relative evaluation. Reference [7] was the first that applied the tournament fitness to IEC. It also proposed a tournament IEC that uses not only which is better of a paired individual but also how better into their final fitness values.

The *tournament1-GA* is a GA in that individuals are evaluated thanks to competitive fitness called *single-elimination tournament* in [13, 14]. Individuals are paired at random, and play one game per pair. Losers of games are eliminated from the tournament… This process continues until the tournament has only one champion left. The fitness of an individual is the number of games played. In the interactive case, i.e. tournament1-IGA, IEC user just chooses one of two displayed individuals.

The *tournament2-GA* is a GA in that individuals are evaluated thanks to another competitive fitness based on single-elimination tournament first proposed in [7]. The fitness is computed based on not only given the number of games played but also how far between a paired individual. For instance, we start by giving a fitness of 10 to the champion of the tournament. Individual which has lost against the champion is given champion's fitness, i.e. 10, minus the difference between it and the champion and so on for all individuals. In the interactive case, the difference between both individuals is supposed to be given by the user and previous fitness values are also made discrete in $n$-evaluation levels.

An individual that has fought against the tournament's champion in the first game, i.e. it lost the tournament at the first game, it will have a better fitness in tournament2 than in tournament1.

### B. Differential Evolution (DE)

The point is that *comparing two vectors* is only the evaluation of DE. We believed that this *paired comparison* is the big advantage of DE to use for IEC. However, work done by [10-12], that seems to be first work on Interactive DE, did not use this big potential to reduce IEC user fatigue and asked an IEC user to choose better individuals among shown all eight individuals.

DE is a population based, stochastic and continuous function optimizer [9] where distance and direction information from current population is used to guide the search process [15]. DE is known to be able to handle non-differentiable, nonlinear and multimodal cost functions, to be parallelized to cope with computation intensive cost functions, easy of use, and well suited for rapid convergence, i.e. consistent convergence to the global minimum in consecutive independent trials.

Basically, for each individual of the population (*parent* or *target vector*), first generate a *mutant vector* by adding weighted difference (*difference vector*) between two randomly chosen vectors (*parameter vectors*) to the third chosen vector (*base vector*). Secondly, the *trial vector* is obtained from the mutant vector and the target vector using binomial or exponential crossover. Finally, target vector is replaced with a better vector of either of the trial one or the target one. There are some variations in how to determine the base vector. More details could be obtained in [9, 15].

## III. EVALUATION TASK

### A. Pseudo-IEC User

Human cannot conduct thousands of evaluations under completely same conditions and is not unreliable to evaluate the convergence of IDE by comparing with those of conventional IEC methods. We should evaluate them with an IEC simulation by designing a pseudo-IEC user even if we evaluate our proposed paired comparison-based IDE with human IEC user later [8].

There are three IEC features that we must realize in the pseudo-IEC user:
  (1) evaluation characteristics with less complexity,
  (2) relative fitness in each generation, and
  (3) discrete fitness in $n$-evaluation levels.
Furthermore, the evaluation characteristics of the pseudo-IEC user should be controlled parametrically.

We realized the (1) using a Gaussian Mixture Model described in section III.B. The (2) is realized thanks to *competitive* fitness function [7, 13, 14] explained in section II.A, and the (3) is realized by dividing the range of the best fitness and the worst fitness obtained thanks Gaussian Mixture Model in $n$-evaluation levels in each generation.

The reason why we use the Gaussian Mixture Models consisting of four Gaussian Mixture functions is to emulate the evaluation landscape in human mind. IEC task may not be a unimodal but not complex because IEC users can reach to satisfactory solutions with less number of population size and generations (see Figure 1).

*B. Gaussian Mixture Model*

Our evaluation tasks are 4 different dimensional Gaussian Mixture Models in 3-D, 5-D, 7-D, and 10-D. We design to make all their function characteristics same to control our experimental complexities by changing only the dimensionality. They are expressed as:

$$f(x_1, \cdots, x_n) = \sum_{i=1}^{k} a_i \exp\left(-\sum_{j=1}^{n} \frac{(x_{ij} - \mu_{ij})^2}{2\sigma_{ij}^2}\right) \quad (1)$$

where $k$ and $n$ are the number of Gaussian functions and the dimensionality, respectively; $k=4$ and $n = 3, 5, 7,$ and 10 in this paper. $a_i$, $\sigma_{ij}$ and $\mu_{ij}$ represent the height, the standard deviation and the central position of the $i$-th Gaussian function, respectively. Parameters values that we used for our model are the following:

$$\sigma = \begin{pmatrix} 1.5 & 1.5 & 1.5 & 1.5 & 1.5 & 1.5 & 1.5 & 1.5 & 1.5 & 1.5 \\ 2 & 2 & 2 & 2 & 2 & 2 & 2 & 2 & 2 & 2 \\ 1 & 1 & 1 & 1 & 1 & 1 & 1 & 1 & 1 & 1 \\ 2 & 2 & 2 & 2 & 2 & 2 & 2 & 2 & 2 & 2 \end{pmatrix}, \quad a = \begin{pmatrix} 3.1 \\ 3.4 \\ 4.1 \\ 3 \end{pmatrix} \text{ and}$$

$$\mu = \begin{pmatrix} -1 & 1.5 & -2 & -2.5 & -1 & 1.5 & -2 & -2.5 & -1 & 1.5 \\ 0 & -2 & 3 & 1 & 0 & -2 & 3 & 1 & 0 & -2 \\ -2.5 & -2 & 1.5 & 3.5 & -2.5 & -2 & 1.5 & 3.5 & -2.5 & -2 \\ -2 & 1 & -1 & 3 & -2 & 1 & -1 & 3 & -2 & 1 \end{pmatrix}.$$ These values have been chosen such as Gaussian Mixture Model landscape has four peaks more or less overlapped to emulate human decision making as shown in Figure 1. In such a case, boundaries between classes of evaluation are not precise.

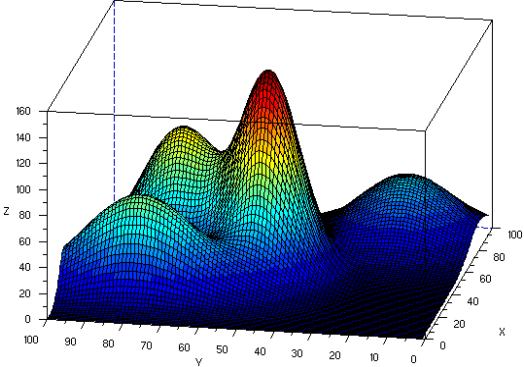

Figure 1. 3D view of a 4D Gaussian Mixture Model used in our experiments.

## IV. EXPERIMENTAL RESULTS

*A. Experimental conditions*

Genotype is a vector of float values; each float value is represented as an array of 12 bits. Vector's dimension is equal to the Gaussian Mixture Model's dimension (3, 5, 7 or 10). Crossover operator is a multipoint crossover with a 100% rate. Mutation rate is 5%. We use a tournament selection operator made of 2 individuals for each tournament. For all experiments, 100 runs are done during 100 generations. For IEC framework, fitness is discretized into 5 evaluation levels (as explained in section III.A).

Our experiments are also conducted with two different population sizes of 16 and 128. The former corresponds to the population size when a human IEC user runs IEC experiments, while the latter corresponds to that for normal EC search and is conducted as the reference for comparing with convergence characteristics of IEC. Population size had also been chosen because of implementation of single-elimination tournament (cf. section II.A); size needs to be *a power of two*, otherwise some individuals will not have the same number of wins.

As mentioned in section II.B, different strategies exist for DE and we used DE/best/1/bin algorithm that is a standard one.

All following remarks are based on the best fitness and not on the average of fitness.

*B. Comparison of DE and three other EC algorithms as References*

In general :
- EC with smaller population size is harder than that with more population size,
- higher dimensional tasks is more difficult,
- tournament GA's use less information for selection than normal GA that selects better parents by comparing all individuals, which implies that the normal GA converges faster than the tournament GA's in general, and
- from the observations of EC research in this decade, DE seems to converges faster than normal GA.

Figure 3 to Figure 10 show all these general observations. This is why we may believe that our four EC algorithms run correctly though we have not applied sign tests to these results.

*C. Comparison of IDE and three other IEC algorithms*

From the practical IEC point of view, we should note that practical generation numbers would be 10 - 20 (maybe at most 30) and a population size would be up to 20. However, convergence curves till 100 generations in graphs are useful to observe general characteristics of IEC.

From convergences till 20-30 generations in Figure 11 to Figure 14, which is a practical use of IEC conditions, it seems that there are no significant differences among 4 EC methods except the simplest task of the 3-D Gaussian Mixture Model. We should apply a sign test or Wilcoxon sign test and confirm whether this view is correct; we do it in the next session.

Comparison of Figure 11 until Figure 14 and Figure 15 until Figure 18 shows that:
- normal IGA with big population size works well in early generations, and
- IDE runs better in later generations, i.e. IDE is a slow starter than IGA. Although we cannot use IEC with big population size and Figure 15 to Figure 18 are not realistic, analysis of the reason of the IDE's slow starter may give us a hit to be applied to practical IEC conditions of less population size in fewer generations and improve IDE in practical conditions.

*D. Results*

Unlike DE and GA, two tournament GA's use rank order fitness, i.e. relative fitness. All IDE, IGA, and tournament IGA use relative fitness. If we make a graph of these relative fitness along with generations, of course we cannot observe convergence. Normalization of fitness also cannot solve it. However, for making comparisons, we must observe their convergence in a searching space with their absolute function values of the individuals evolved based on the relative fitness.

## V. DISCUSSION

We statistically tested whether DE or IDE is significantly better than others or worse than the best at each generation. Results are shown in Figure 2.

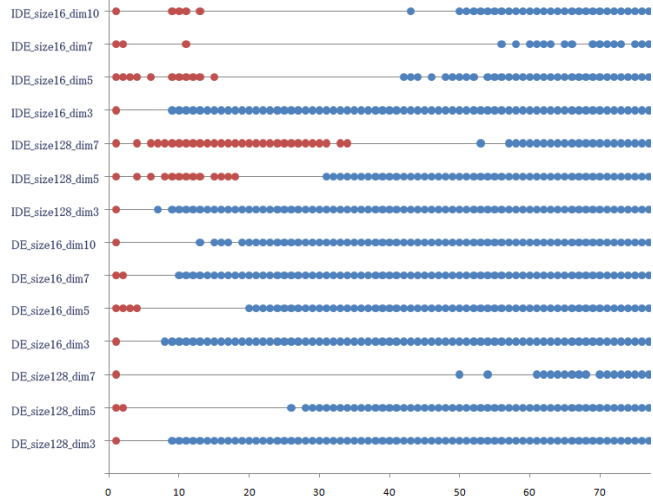

Figure 2. Sign test for DE (IDE); blue dots mean that DE (or IDE) is significantly better than others in the (1) case, and the red dots mean that DE (IDE) is significantly poorer than at least one of others.

Our observation is that
- DE is always the best or at least not poorer than others significantly in early generations. It becomes the best after 10-20 generations for three different complexities of tasks with small (16) and large (128) population size.
- IDE with 16 population size shows similar tendency. However, the generations that the IDE becomes significantly better than other after about the 10th generation for simple task (3-D Gaussian mixture model) and 40th generation for complex tasks (5-D, 7-D, and 10-D models).
- IDE with 128 population size does not have reality as IEC, and its results themselves are not important.

## VI. CONCLUSION

Better IEC user interface and fast convergence are necessary to reduce IEC user fatigue. We proposed paired comparison-based IDE that can reduce IEC user fatigue greatly than comparison of all individuals. Especially, it is effective when IEC tasks handle individuals displayed time-sequentially, i.e. sounds or movies. Since this advantage is obvious, we evaluated another key point, convergence speed and show the superiority of IDE to IGA and two tournament IGA's. From these two advantages of the IDE, we can say that our proposed IDE is the better than conventional IDE, IGA, and tournament IGA's.

Our next step of this research is to evaluate these advantages of our IDE found though IEC simulations is really effective for human IEC. We are planning to evaluate the proposed IDE using real human users.

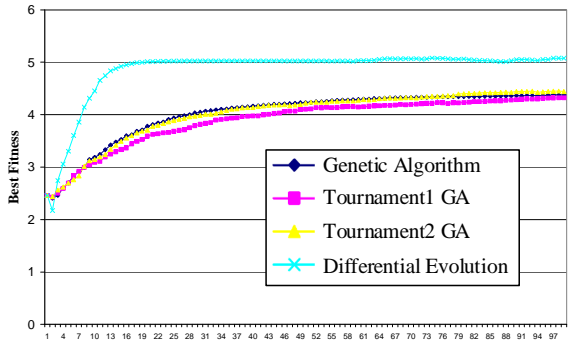

Figure 3. EC Task is 3-D Gaussian Mixture Model (16 individuals).

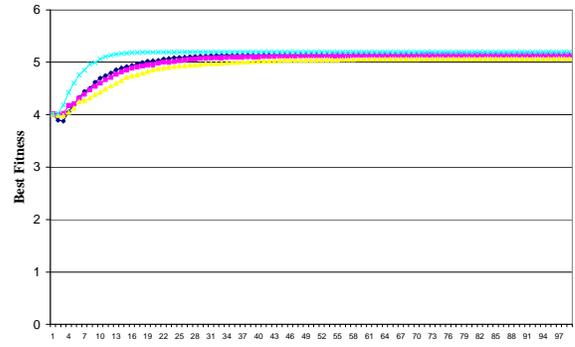

Figure 7. EC Task is 3-D Gaussian Mixture Model (128 individuals).

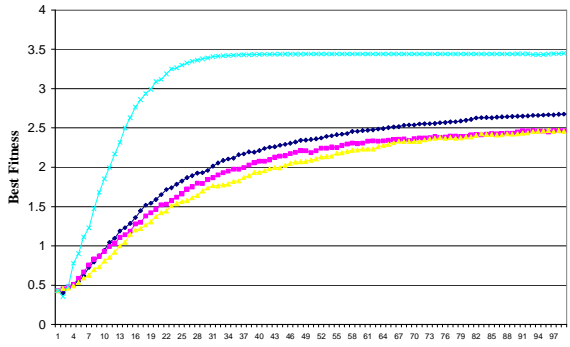

Figure 4. EC Task is 5-D Gaussian Mixture Model (16 individuals).

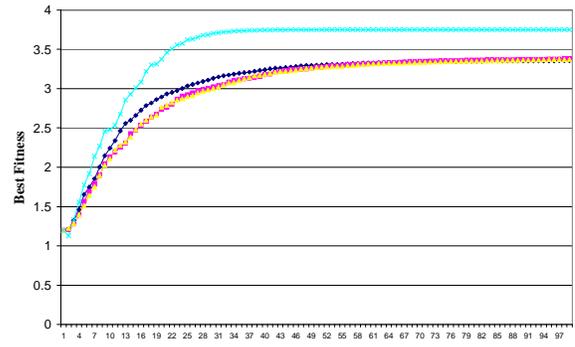

Figure 8. EC Task is 5-D Gaussian Mixture Model (128 individuals).

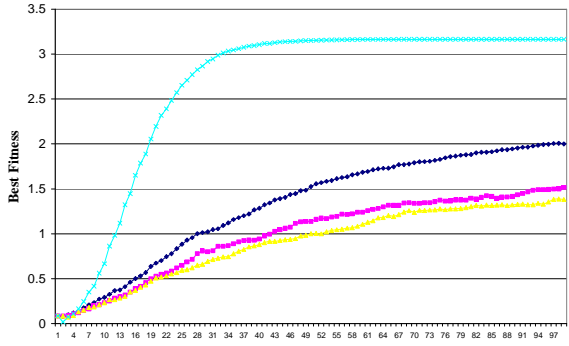

Figure 5. EC Task is 7-D Gaussian Mixture Model (16 individuals).

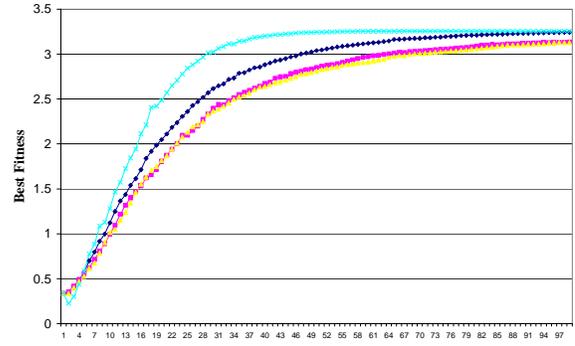

Figure 9. EC Task is 7-D Gaussian Mixture Model (128 individuals).

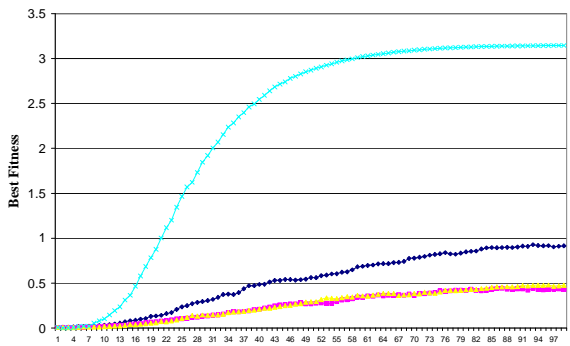

Figure 6. EC Task is 10-D Gaussian Mixture Model (16 individuals).

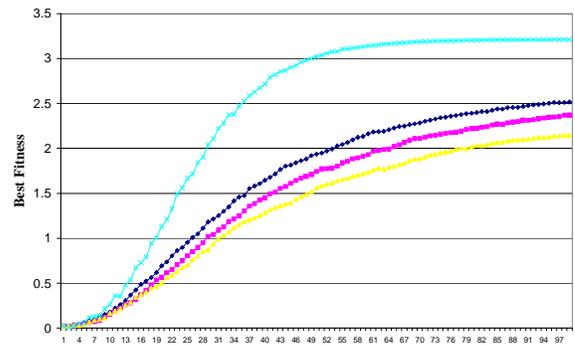

Figure 10. EC Task is 10-D Gaussian Mixture Model (128 individuals).

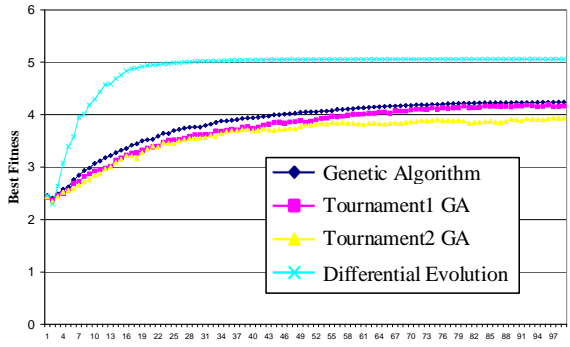

Figure 11. IEC Task is 3-D Gaussian Mixture Model (16 individuals).

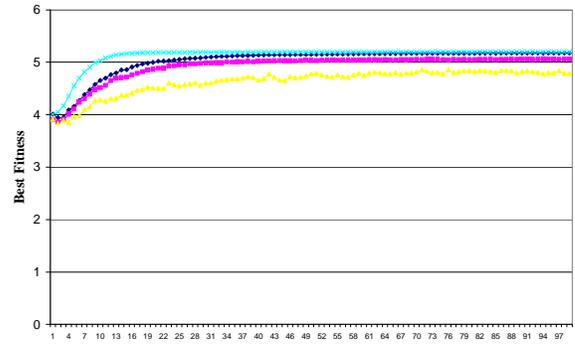

Figure 15. IEC Task is 3-D Gaussian Mixture Model (128 individuals).

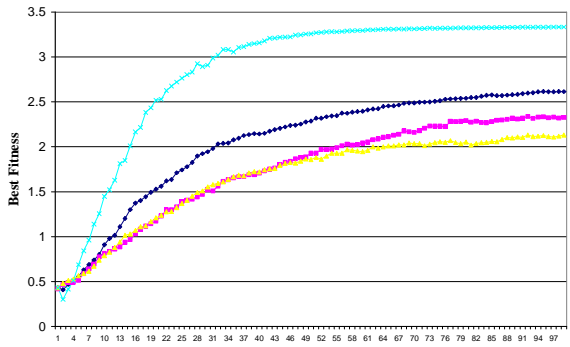

Figure 12. IEC Task is 5-D Gaussian Mixture Model (16 individuals).

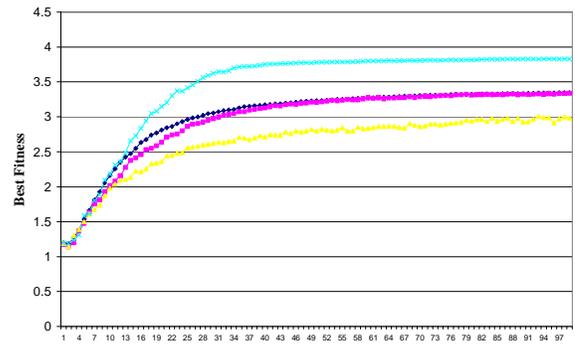

Figure 16. IEC Task is 5-D Gaussian Mixture Model (128 individuals).

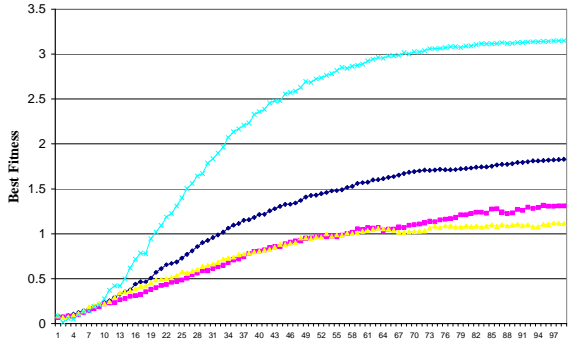

Figure 13. IEC Task is 7-D Gaussian Mixture Model (16 individuals).

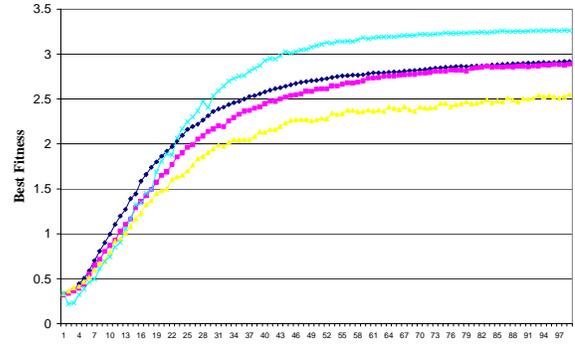

Figure 17. IEC Task is 7-D Gaussian Mixture Model (128 individuals).

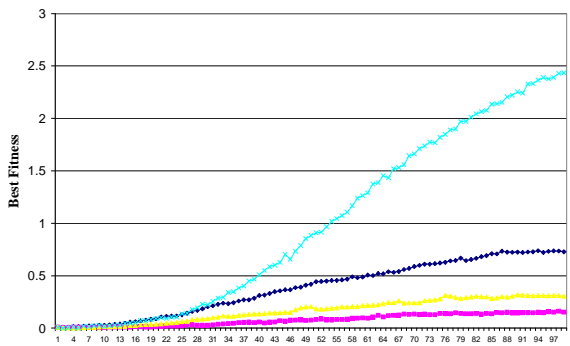

Figure 14. IEC Task is 10-D Gaussian Mixture Model (16 individuals).

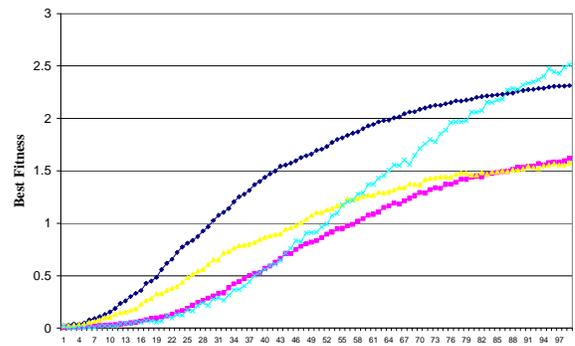

Figure 18. IEC Task is 10-D Gaussian Mixture Model (128 individuals).